\documentclass[letterpaper, 10 pt, conference]{ieeeconf}  

\IEEEoverridecommandlockouts                              
\usepackage{multicol}
\usepackage[bookmarks=true]{hyperref}
\usepackage{cite}
\usepackage{amsfonts}
\usepackage{amsmath}
\usepackage{graphicx}
\usepackage{pgfplots}
\usepackage{tikz}
\usepackage{subcaption}
\usepackage{hyperref}
\usepackage{algorithm}
\usepackage{algpseudocode}
\usepackage{float}
\usepackage{todonotes}
\usepackage{amsmath}

\usepackage{pgfplots}
\usepackage{pgfplotstable}
\pgfplotsset{compat=newest}
\colorlet{negro}{black}
\colorlet{gris}{black!70}
\colorlet{rojo}{red!70!black}
\colorlet{rojol}{red}
\usepackage{booktabs}
\usepackage{color, colortbl}
\definecolor{Gray}{gray}{0.9}
\definecolor{LightCyan}{rgb}{0.88,1,1}
\newcolumntype{a}{>{\columncolor{Gray}}c}
\newcolumntype{b}{>{\columncolor{white}}c}
\usepackage{epstopdf}
\newcommand{\ts}{\textsuperscript}

\usepackage{optidef}
\usepackage{booktabs}

\title{\LARGE \bf
A Pedestrian Detection and Tracking Framework for Autonomous Cars: Efficient Fusion of Camera and LiDAR Data 
}

\author{Muhammad Mobaidul Islam, Abdullah Al Redwan Newaz, and Ali Karimoddini\ts{*}
\thanks{M. M. Islam, A. A. Redwan Newaz,  and A. Karimoddini are with the Department of  Electrical and Computer Engineering, 
North Carolina A\&T State University, Greensboro, NC 27411 USA.}%
\thanks{\ts{*}Corresponding author: A. Karimoddini. Address: 1601 East Market Street,
Department of Electrical and Computer Engineering
North Carolina A\&T State University
Greensboro, NC, US 27411.  Email: {\tt\small akarimod@ncat.edu}
(Tel: +13362853313).}%
}

\begin{document}
\maketitle
\begin{abstract}
This paper presents a novel method for pedestrian detection and tracking by fusing camera and LiDAR sensor data. To deal with the challenges associated with the autonomous driving scenarios, an integrated tracking and detection framework is proposed. The detection phase is performed by converting LiDAR streams to computationally tractable depth images, and then, a deep neural network is developed to identify pedestrian candidates both in RGB and depth images. To provide accurate information, the detection phase is further enhanced by fusing multi-modal sensor information using the Kalman filter. The tracking phase is a combination of the Kalman filter prediction and an optical flow algorithm to track multiple pedestrians in a scene. We evaluate our framework on a real public driving dataset. Experimental results demonstrate that the proposed method achieves significant performance improvement over a baseline method that solely uses image-based pedestrian detection.

\end{abstract}

\section{INTRODUCTION}

Real-time accurate pedestrian detection and tracking are crucial to ensure safety and reliability in autonomous driving \cite{islam2020pedestrian}.  This often requires the estimation of multiple pedestrian trajectories from multi-modal sensor data. There are, however, several challenges that make pedestrian detection and tracking a notoriously hard task. For instance, pedestrians may appear in a scene with different articulations of body parts, partial occlusion, and appearance in various poses. Furthermore, the detection and tracking may fail due to the sensitivity of some sensors under low illumination or bright sunlight conditions. Despite all these challenges, it is vitally important for autonomous driving to achieve accurate and robust pedestrian detection and tracking under various conditions in real-time to be able to transition from the innovation space to real-world operations \cite{homaifar2019operationalizing,rabby2019review}. 

Pedestrian tracking consists of two main stages:
pedestrian detection and association of the detected pedestrians with current and past estimations. Generally, there are two streams of research directions for pedestrian tracking \textemdash either use of deep learning models in an end-to-end fashion or the development of a machine learning pipeline with a mix of deep learning and classical methods, to enable pedestrian tracking. Some of the  notable end-to-end deep learning-based pedestrian tracking methods include TrackR-CNN~\cite{voigtlaender2019mots}, Tracktor++~\cite{bergmann2019tracking} and JDE~\cite{wang2019towards}. These methods are created by augmenting existing pedestrian/object detectors, e.g., by adding additional recurrent layers to the detection head to incorporate the temporal context of the scenes. 
There are few end-to-end multimodal sensor fusion mechanisms that use 3D LiDAR and camera data for pedestrian tracking~\cite{milan2017online,frossard2018end}.
These methods commonly suffer from low runtime efficiency because of the complex network architecture. 
On the contrary, machine learning pipeline based methods combine a pedestrian detector with the classical filtering technique to achieve better runtime efficiency~\cite{asvadi20163d,wojke2017simple,wang2019towards}. 
These methods mostly rely on a specific sensor modality such as camera data. Therefore, these methods suffer from either rich feature descriptors or sensitivity to environmental conditions (e.g., illumination variations in case of using a camera), and hence, these methods are not robust enough for detecting and tracking pedestrians.
\begin{figure}
    \vspace{3pt}
    \centering
    \includegraphics[height=3cm,width=8.5cm]{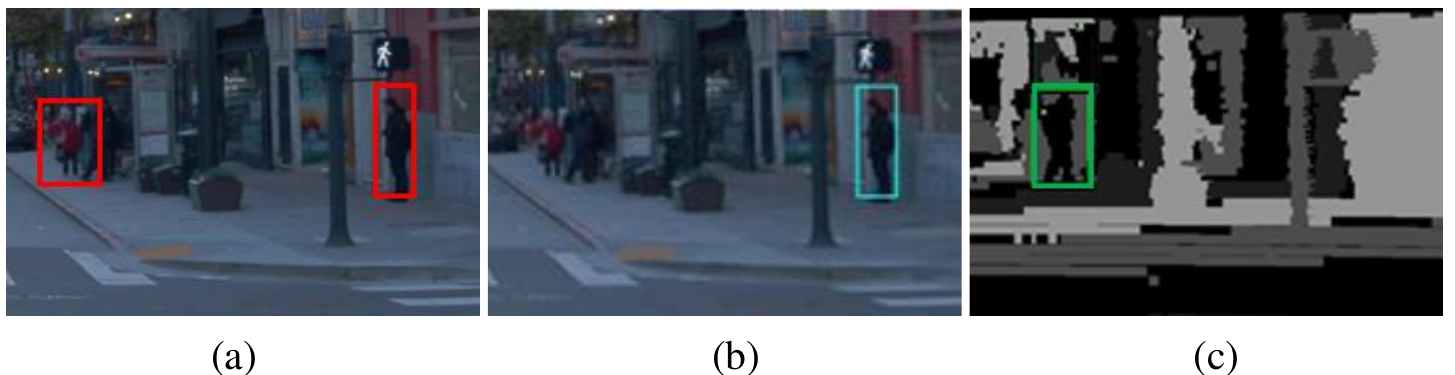}
    \caption{\small{Fusion of multi-modal sensor data can improve pedestrian detection and tracking accuracy: (a) ground-truth bounding boxes of two anticipated pedestrians, (b) detector can detect only one pedestrian on camera image, (c) and another pedestrian on corresponding depth image from LiDAR scan. Therefore, the best result can be achieved by combining these independent results from (b) and (c) into a joint prediction} }
    \label{fig:Motivational_image}
    \vspace{-10pt}
\end{figure}

In this work, we hypothesize that a fusion between camera and LiDAR can enhance the robustness and accuracy of pedestrian detection and tracking.
Fig.~\ref{fig:Motivational_image} shows a realistic event where it is anticipated from a pedestrian detector to detect at least two pedestrians. As it is obvious from Fig.~\ref{fig:Motivational_image}.b, the detector can detect only one pedestrian on the camera image. On the other hand, as shown in Fig.~\ref{fig:Motivational_image}.c,  another pedestrian can be detected on the corresponding depth image from the LiDAR scan.
Therefore, in this paper, we propose a fusion framework that combines LiDAR streams and camera data as well as the estimation of vehicle motion using the camera-based optical flow method. To achieve computationally tractable and real-time pedestrian tracking, first,  LiDAR 3D streams are converted to 2D depth images and then are fed to a pedestrian detector by vertically concatenating these frames with camera images. Finally, the Kalman filter is used for fusing the prediction over the concatenated images as well as the locations of previously detected pedestrians in the current frame.


In summary, the contributions of this paper are as follows:
\begin{enumerate}
        \item we develop a method for real-time sensor fusion of camera and LiDAR data,
        \item we develop a real-time accurate pedestrian detection and tracking framework, 
     \item we also integrate optical flow information into the developed tracking technique for achieving accurate predictions of multiple pedestrians over the RGB and depth images.
\end{enumerate}
The organization of the paper is as follows. Section~\ref{related} reviews related works. Section~\ref{sec:Prelim} provides the preliminaries about pedestrian detection. Section~\ref{Proposed} describes the proposed fusion architecture for tracking multiple pedestrians. Section~\ref{Experiment} provides the experimental results for the evaluation of the proposed method. Finally, Section~\ref{Conclusion} provides the conclusion remarks. 

\section{Related work}\label{related}

In recent years,  numerous approaches for detecting and tracking pedestrians in sequential images have grown steadily. With the recent advancement in deep learning, we can utilize machine learning models to accurately detect and classify pedestrians in complex scenarios. In this section, we will begin with a brief overview of learning-based pedestrian detection, then some existing fusion techniques for combining multimodal sensor data, and finally a brief overview of pedestrian tracking.  
 
\subsection{Learning-based Pedestrian Detection}
Pedestrian detection from RGB images is an important yet difficult task. Recent works focus on improving the robustness and accuracy using deep neural networks~\cite{tian2015deep,zhao2016faster,zhang2018ssd,zadobrischi2020pedestrian,zhang2018occluded}. Though these methods exhibit satisfactory performance in well-lit environments, they struggle to detect pedestrians in low light conditions such as nighttime, dawn, sunrises, and sunsets. This is because it is hard to generate shape information from images in ill-lit environments.

On the other hand, the LiDAR can provide comparatively better shape features under these scenarios. LiDAR-based feature extraction for pedestrian detection is studied in early research~\cite{premebida2009exploiting}. As LiDAR can provide the only geometric feature of pedestrians, inferring context-aware relations of pedestrians' body parts is one way to distinguish among multiple pedestrians in a complex scenario~\cite{oliveira2010context}. 
While using the LiDAR sensor data \textemdash the distance, intensity, and width of the received pulse signal in each scanning direction, pedestrians can be classified using a clustering algorithm~\cite{ogawa2011pedestrian}.
To improve the classification performances of pedestrians, hand-crafted features such as slice feature and distribution of reflection intensities are explored~\cite{kidono2011pedestrian}. The slice provides human body information based on body height and width ratio. Some works are focused on the density enhancement method for improving the sparse point cloud of LiDAR and they provide an improved shape feature for long-distance pedestrian detection~\cite{li2015density,liu2019pedestrian}.
Pioneering work on the conversion of a 3D point cloud of LiDAR into the 2D plane extracts both hand-crafted features and learned features, and then trains a support vector machine (SVM) classifier to detect pedestrians~\cite{lin2018pedestrian}.  
Later, 3D point clouds are converted into 2D panoramic depth maps and these depth maps are used in pedestrian detection~\cite{chen2020pedestrian}.
Even though the LiDAR provides better results in the nighttime while it is difficult to get shape features using the camera or compare to a distorted image frame, camera-based methods  perform better for long-distance pedestrians in the daytime where they appear in small sizes. 
The best result can be achieved by fusing both of these sensors to jointly predict pedestrians.

\subsection{Pedestrian Detection by Fusing Sensors}
Since using the LiDAR or the camera independently unveils their own limitations, it becomes an interesting research direction to fuse different sensor modalities. In this setting, the improvement can be achieved from the use of multiple views of the pedestrian by learning a strong classifier that accommodates both different 3D points of view and multiple flexible articulations.
In order to integrate multiple sensor modalities, several fusion mechanisms are investigated~\cite{premebida2009lidar,premebida2013fusing,premebida2014pedestrian,gonzalez2015multiview,schlosser2016fusing,matti2017combining,kim2018pedestrian}. These sensor fusion techniques mostly focus on either combining feature information from different sensors or generating candidate regions from one sensor and map these candidate regions to other sensor information. 
For instance, a deformable part detector is trained using optical images and depth images generated from 3D point clouds using upsampling technique~\cite{premebida2014pedestrian}.
Some fusion techniques cluster the LiDAR point cloud to generate candidate regions and map these regions on an image frame for detecting pedestrians~\cite{Lahmyed2016,matti2017combining}. Most of these methods sacrifice runtime performance while improving detection accuracy. Therefore, a balanced fusion mechanism is needed to deal with the trade-off between accuracy and speed. 

\begin{figure*}
    \vspace{3pt}
    \centering
    \includegraphics[height=6cm]{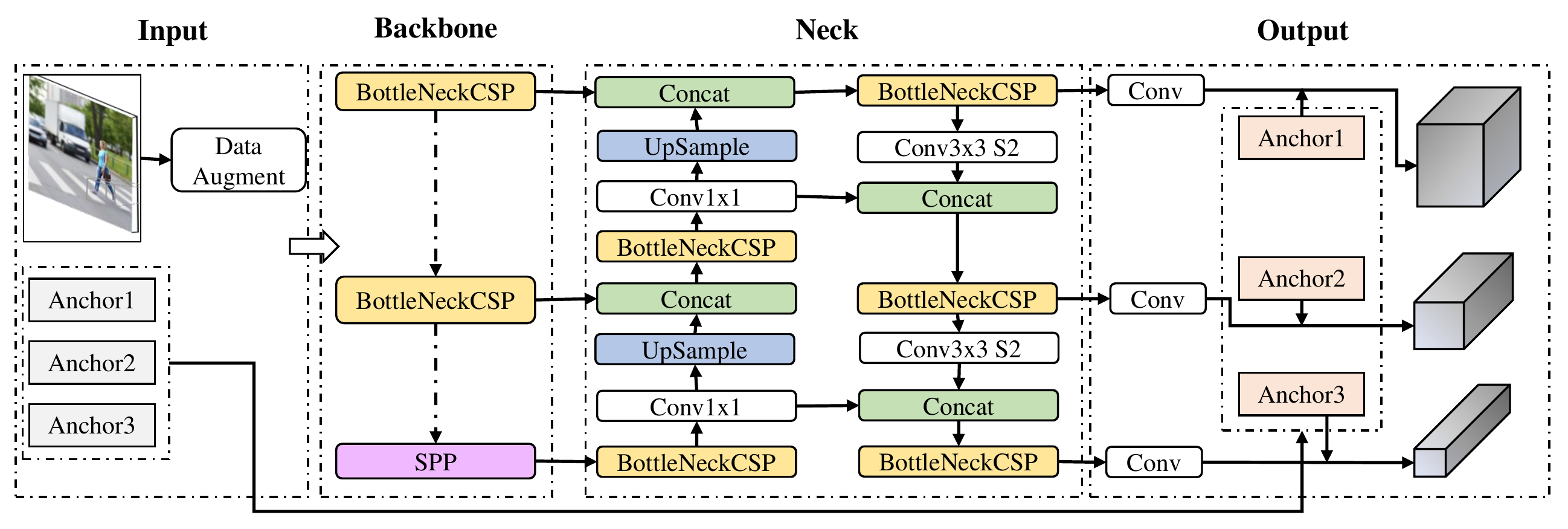}
    \caption{\small{\textbf{Architecture of YOLOv5:} YOLOv5 has four major sections. It starts with input then merges to the backbone for feature extraction. Feature information is enhanced in the neck section, and the output section produces a bounding box with confidence scores for a specific class of objects}}
    \label{fig:Architecture_of_Yolo}
    \vspace{-10pt}
\end{figure*}
\subsection{Pedestrian Tracking}
Recent pedestrian trackers are designed mostly based on end-to-end deep learning networks. 
A common approach is adding recurrent layers with the detector module. For example, the ROLO~\cite{Ning2017ROLO} is the combination of the convolutional layers of YOLO and the recurrent unit of LSTM. 
TrackR-CNN~\cite{voigtlaender2019mots} is considered as a baseline method of multi-object tracking that adds instance segmentation along with multi-object tracking. Tractor++ is an efficient multiple object tracking that utilizes the bounding box regression on predicting the position of an object in the next frame where there is no train or optimization on tracking data~\cite{bergmann2019tracking}. Besides, a single object tracking along with semi-supervised video object segmentation based on siamese neural network is introduced in~\cite{wang2019fast}. 
On the other hand, the machine learning pipeline-based methods such as the Deep SORT which integrates appearance information along with Simple Online and Realtime Tracking (SORT) technique, adopts a single hypothesis tracking methodology with the recursive Kalman filter and the frame-by-frame data association. This technique focuses on an offline pre-training stage where the model learns a deep association metric on a large-scale person re-identification dataset~\cite{wojke2017simple}. 
A single-stage efficient multi-object tracking is introduced in~\cite{wang2019towards}, where target detection and appearance are embedding to be learned in a shared way, and a Kalman filter is used for predicting the locations of previously detected objects in the current frame. 
While considering the LiDAR data for pedestrian tracking, a stochastic optimization method is introduced in~\cite{granstrom2017pedestrian} that merges the clustering and assignment task in a single stage.

Inspired by these works, we use both LiDAR and camera sensors to complement individual sensor limitations on detection and tracking performances. Thus, our solution can be applied to a wide variety of complex scenarios. 
\section{Preliminaries}\label{sec:Prelim}

Pedestrian tracking is a challenging problem as pedestrians need to be firstly detected in the current frame and then associated frames. The success in deriving a good tracker is mainly governed by a superior detector.
In this work, we use YOLOv5~\cite{glenn_jocher_2021_4679653} as a base module of the pedestrian detector. Therefore, we will briefly explain the working principle of YOLOv5 as a pedestrian detector.
Since the YOLOv5 architecture is the same as the YOLOv4~\cite{bochkovskiy2020yolov4} except the training procedure, we will describe the architecture of the YOLOv4 and then will highlight the training differences.


Fig.~\ref{fig:Architecture_of_Yolo} illustrates the architecture of YOLOv4 which can be segmented into four major parts: input, backbone, neck, and output. In the input section, the network takes an image and completes a data augmentation procedure that uses a data loader for scaling, color space adjustments, and mosaic augmentation. Among these augmentation techniques, mosaic augmentation firstly introduces in YOLOv4. 
The mosaic augmentation combines four training images into one in certain ratios to simulate four random crops which help to detect small-scale and partially occluded pedestrians. 

After data augmentation, the augmented image is feed into the backbone of the network.
In the backbone section, a BottleNeckCSP is used which is a modification of DenseNet~\cite{huang2017densely}. Using BottleNeckCSP different shallow features like edges, colors, etc., are extracted. During training, the backbone module learns these features.
Besides, an additional Spatial Pyramid Pooling (SPP) block is used to increase the receptive field and separate the most important features from the feature maps of the BottleNeckCSP.
The next part of the network is the neck part where the network enhances the understanding and extraction of the shallow features adopted in the backbone part. 
To do that a Path Aggregation Network (PANet) is used that includes a bottom-up augmentation path in conjunction with the top-down path used in Feature Pyramid Network (FPN). The PANet processes combine and analyzes the extracted features and finally optimizes based on the target of the model.
The last part of the network is output where the model yields the detection results using dense predictions. Dense predictions provide 
a vector by combining predicted bounding boxes and confidence scores for the classified pedestrians.

During the training process of YOLOv5, the floating-point precision is set to 16 bit instead of the 32-bit precision used in YOLOv4. Therefore, YOLOv5 exhibits higher performances than Yolov4 under certain circumstances. 

\section{Proposed Methodology}\label{Proposed}
In this section, we will explain our proposed methodology. At the high level, we propose Fused-YOLO which is an integrated framework of multi-modal sensor information to track pedestrians at a real-time speed. At the core of our proposed solution lies the distinction between improving detection accuracy and limiting the computational complexity.

\begin{figure*}[t]
    \vspace{3pt}
    \centering
    \includegraphics[height=6.0cm]{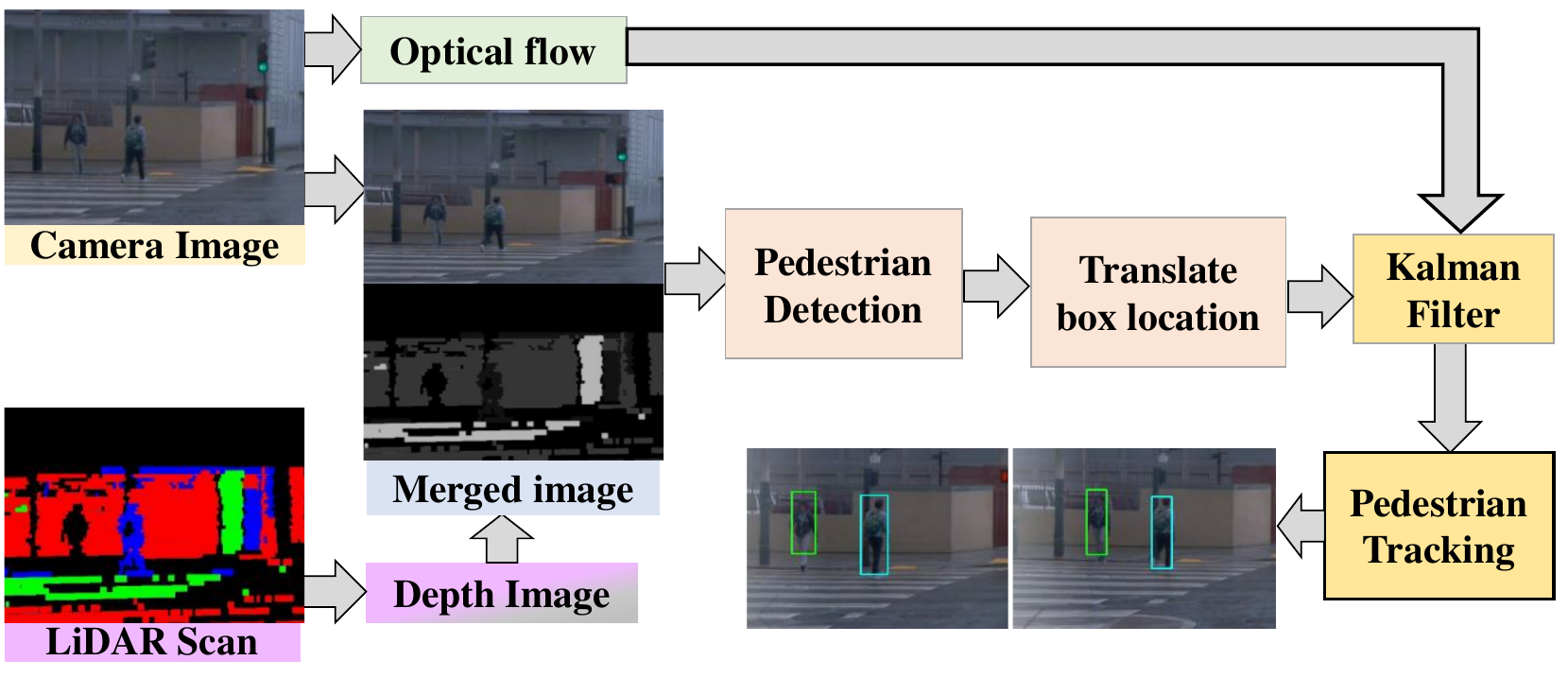}
    \caption{\small{\textbf{The proposed pedestrian detection and tracking framework:} A camera image and its synchronized LiDAR scan are the input of this framework. First, the LiDAR scan is converted into a depth image and then merge with the camera image. Next, merged fed into pedestrian detection module and outputs are mapped into the single-camera image frame by translating the detected boxes locations. Finally, for pedestrian tracking, a Kalman filter is implemented in which inputs are these translated bounding boxes and optical flow of input image.}}
    \label{fig:Architecture_of_proposed_fusion_mechanism}
    \vspace{-10pt}
\end{figure*}

\subsection{Conversion from LiDAR Scans to Depth Images}

Camera-based pedestrian detection systems suffer from either low illumination or over-exposed images. 
It is a better idea to complement the system with a LiDAR which acts as the primary depth sensor due to its high accuracy and long sensing range. LiDAR scan produces sparse point clouds, albeit this representation of data is rather challenging to incorporate as an input to neural networks. Instead, depth images are better correspondents of point clouds that are easy to manipulate constructively. Therefore, we convert a 3D LiDAR scan to the depth image in 2D image space.

Formally, the LiDAR stream consists of a sequence of registered 3D scans $\{S_1 , S_2 , . . ., S_t \}$ arriving at time points ${t_1, t_2, . . ., t_t}$. Each scan $S_t$ is a point cloud, i.e., a set of 3D points, $S_t = \{\mathbf{p}_1,\mathbf{p}_2, . . . , \mathbf{p}_i\}$ and $\mathbf{p}_i:=\{x,y,z\}$ represents the Euclidean coordinate.  Due to the huge amount of memories that are required over time, it is inefficient to work upon the raw point clouds. Instead, we can convert the 3D scan to a 2D depth map. A depth image can be thought of as a 2D grid map comprised of $u_n$ cells. To generate the depth image we need to compute the distance of the scan objects from a viewpoint in such a way that maps $\mathbf{p}_i$ to $u$. Then, we transform each point in point clouds from the Euclidean coordinate $(x,y,z)$ to Spherical coordinate $(\theta,\phi, r)$. This way we can map each point to the corresponding grid cell such that $ u:\{\theta,\phi\}\to r$. The pixel values of depth images lie in either gray or RGB color spaces. For the grayscale image, we normalize each cell value in the grid map to $0 \to 255$ to the known maximum depth value and thus the intensity of the gray image represents the depth information. On the other hand, For the RGB scale image, we assign a distinct color from the RGB space to each cell value in the grid map based on the $r$ parameter.

\subsection{Fusion with Depth Image}
Our goal is to predict pedestrians in a joint space that combines both the RGB and the depth spaces. Although detecting pedestrians in RGB images is a common practice, there are few ways to incorporate depth images to detect pedestrians in a joint space. An end-to-end deep learning network takes an RGB image and corresponding depth image as input to generate joint predictions over pedestrians. In another setting, an RGB image and a depth image can be processed sequentially using a single network. However, in the former case, the network architecture becomes very complex to be able to process depth and RGB images in an end-to-end fashion. In the later case, the network requires to process sequential call which causes a huge runtime overhead in a long run.    

Our solution utilizes the Kalman filter along with parallel processing of RGB and depth images. To predict in a joint space, first, we project the LiDAR scan as a depth image to the RGB camera space. Let $x_{r}$ and $x_{d}$ be the RGB and depth images, respectively. 
Since (in common settings) the positions of camera and LiDAR are fixed but the resolution of camera image and the depth image from the LiDAR scan varies in size, we can project the depth image to the RGB camera space with either zero padding to the smaller image or cropping each of them into a same size. We denote this synchronized depth image by $x_{s}$. 
Second, we vertically concatenate the RGB image  $x_{r}$ and the zero-padded depth image $x_f$ by resizing each of them into a fixed size such that $\mathbf{x} = \{x_s, x_r\}$. Although it is possible to use an image classification network to predict pedestrians directly over $x_s$, it requires multiple calls for the joint prediction, i.e., the $x_{r}$ and the $x_{s}$ need to be fed to the pedestrian detection and pedestrian classification models, respectively. Thus, concatenating the $x_{r}$ and the $x_{s}$ into $\mathbf{x}$ reduces the number of calls to different models and significantly improves the runtime efficiency.
Finally, we feed this concatenated image $\mathbf{x}$ to the pedestrian detector $f$ to obtain bounding boxes and scores over fused images such that $\hat{y} = f(\mathbf{x})$. 
Form the prediction $\hat{y}$, we can also separate individual predictions the $\hat{y}_r$ and the $\hat{y}_s$ for the $x_r$ and the $x_s$ correspondents, respectively. Since we vertically concatenate the $x_{s}$ with the $x_{r}$, we can calculate an offset $o$ based on the height of $x_r$. Then, we translate each bounding box $b_s \in \hat{y}_s$ to the down by $o$.  

Overlaying $\hat{y}_s$ to $\hat{y}_r$ may raise three distinct types of scenarios. Firstly, $\hat{y}_s$ reduces the miss detection by accurately detecting pedestrians. Secondly, $\hat{y}_s$ provides redundant inference with respect to $\hat{y}_r$. Finally, $\hat{y}_s$ does not improve detection accuracy since it cannot detect any pedestrians. To overcome these scenarios, we utilize a Kalman filter to evaluate the joint predictions systematically. In our next subsection, we will describe the proposed Kalman filter in detail.

\subsection{Integrated Framework for Pedestrian Tracking}
The Kalman filter has been extensively applied in pedestrian tracking from the camera stream. Our framework uses such a technique to predict and update the pedestrian trajectories from the continuous camera and LiDAR streams. Our integrated framework augments the capability of the existing pedestrian tracking method by fusing depth information. To track multiple pedestrians in a frame, our framework uses three important information, i.e., bounding boxes from the RGB images, optical flow between consecutive RGB image frames, and bounding boxes from the depth images.   

One of the important properties of the Kalman filter is that the state vector is a hidden parameter and the observation provides useful information to update the state vector. Therefore, in our setting while using the Kalman filter, the observations, i.e., bounding boxes, from the detector are not directly useful for tracking pedestrians. 
Basically, the proposed Kalman filter-based tracking has two stages: the \textit{prediction} and the \textit{update} stages. In the prediction stage, the bounding boxes for pedestrians are predicted using the corresponding state of the bounding boxes in the previous frames. In the update stage, the observation of pedestrians in the current frame is used to update the predicted states of pedestrians.  

\noindent Let $\mathbf{s}^i_t$ be the state vector of $i$\ts{th} bounding pedestrian window in frame $t$. To track multiple pedestrians, it is convenience to have multiple Kalman
filters, e.g., one for each pedestrian detected in the frame as follows  
\begin{equation*}
    \begin{split}
        &\mathbf{s}^i_t = \mathbf{F}^i_{t-1}\mathbf{s}^i_{t-1} + w_{t-1}, \\ 
        &\mathbf{z}_t^i = \mathbf{H}^i_t \mathbf{s}^i_t + v_t,
    \end{split}
\end{equation*}
where $\mathbf{F}^i_{t-1}$  and $\mathbf{H}^i_t$ denote the state transition and the
measurement matrices for the $i$\ts{th} pedestrian, respectively. The vectors $w_{t-1}$ and $v_t$ are noise terms which are assumed to be Gaussians with zero mean and covariance matrices $Q_t$ and $R_t$.
The prediction stage involves reasoning about the state vectors and their associated error covariance matrices at time $t$ given the measurements up to $t-1$ as follows:
\begin{equation*}
    \begin{split}
        &\mathbf{s}^i_{t|t-1} = \mathbf{F}^i_{t-1}\mathbf{s}^i_{t-1} + w_{t-1}, \\ 
        &\mathbf{P}^i_{t|t-1} = \mathbf{F} \mathbf{P}^i_{t-1|t-1} \mathbf{F}^T + Q_t.
    \end{split}
\end{equation*}
Next, in the update stage updates, the state vectors and their error covariance matrices with the current observations are as follows:
\begin{equation*}
    \begin{split}
&\mathbf{K}_t^i =  \mathbf{P}^i_{t|t-1} \mathbf{H}^T \left(\mathbf{H} \mathbf{P}^i_{t|t-1} \mathbf{H}^T + R_t\right)^{-1},\\ 
&\mathbf{s}^i_{t|t} = \mathbf{s}^i_{t|t-1} + \mathbf{K}_t^i \left(\mathbf{z}_t^i - \mathbf{s}^i_{t|t-1} \right),\\
&\mathbf{P}^i_{t|t-1} = \left(I - \mathbf{K}_t^i \mathbf{H} \right)\mathbf{P}^i_{t|t-1},
    \end{split}
\end{equation*}
where $\mathbf{K}_t^i$ is the Kalman gain which emphasizes how prediction and measurement are intimately related.
Therefore, the process of fusion begins with identifying observation models and associated measurement noises for each observation modality. For instance, in our settings, the bounding boxes from the RGB and depth images are considered as positional information whereas the optical flow provides the velocity information only. This way we can assign a separate observation model for updating the joint prediction state.

\section{Experiment results}\label{Experiment}
We evaluate the pedestrian detection performances in terms of Miss Rate (MR) vs False Positive Per Image (FPPI) and also provide accuracy, precision, recall, and run-time efficiency of the model. We conduct our experiments on a 64-bit Ubuntu 18.04  server that has an  Intel(R)  Core(TM)  i9-7900XCPU @ 3.30GHz with 64GB memory. In our setup, we also have an NVIDIA GeForce RTX 2080 GPU with memory.
\subsection{Dataset}
We perform our experiments on the Waymo open dataset which contains a wide range of diverse examples. Waymo data are collected among Phoenix, Mountain View, and San Francisco cities in the USA, plus it contains daytime and nighttime driving data~\cite{sun2020scalability}. This dataset is recently released and comprising of large-scale multimodal sensor data, i.e., high-resolution camera and LiDAR data. In particular, the dataset is collected using five LiDAR sensors and five high-resolution pinhole cameras and contains four object classes: Vehicles, Pedestrians, Cyclists, Signs. 
It has 12.6M high-quality 3D bounding box labels in total for 1,200 segments for LiDAR data.
On the other hand, it has 11.8M 2D tightly fitting bounding box labels in total for 1,000 segments of camera data. In our setup, we use front camera images and project LiDAR data onto their corresponding camera images.
 
\begin{figure}[htb]
    \centering
    \includegraphics[width= 8cm]{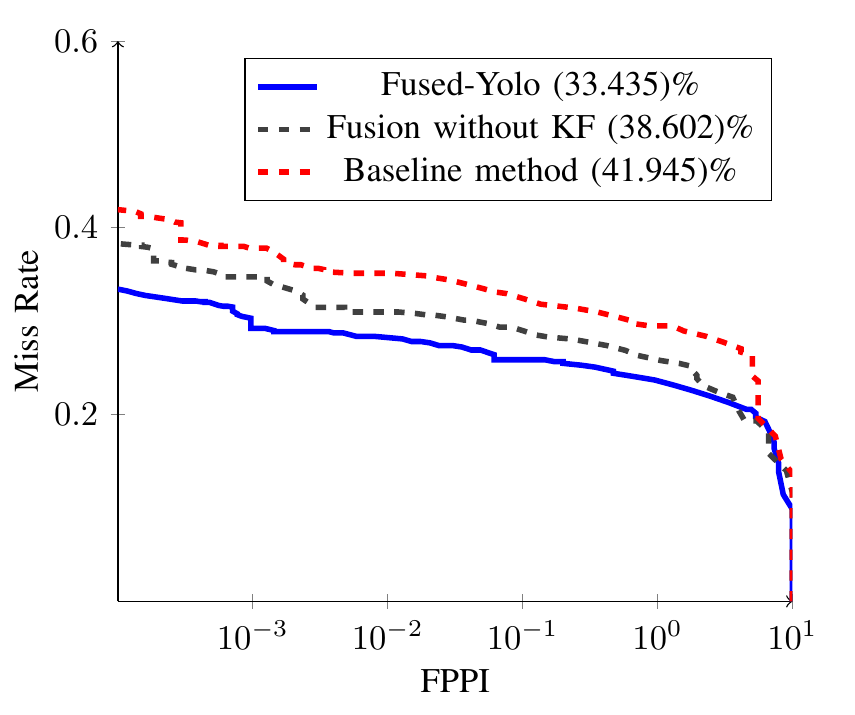}
        \caption{\small{The comparison among Fused-YOLO, Fusion without applying Kalman Filter, and Baseline YOLOv5. Fused-YOLO model get the lowest miss rate of $33.435\%$ among others.}}
    \label{fig:Miss_rate_vs_fppi}
\end{figure}

\begin{table}[htb]
\centering
\begin{tabular}{@{}lllllll@{}}
\toprule
Model & FP & TP  & FN & Accuracy & Precision & Recall \\ \midrule
YOLOv5   & 12 & 105 & 75 & 0.546    & 0.897     & 0.583  \\
Fused-YOLO & 12 & 117 & 63 & 0.609    & 0.900      & 0.650   \\ \bottomrule
\end{tabular}
\caption{Evaluation Metrics}
\label{tab:evaluation_metrics}
\end{table}

\begin{figure*}
    \vspace{4pt}
    \centering
    \includegraphics[height=4.8cm]{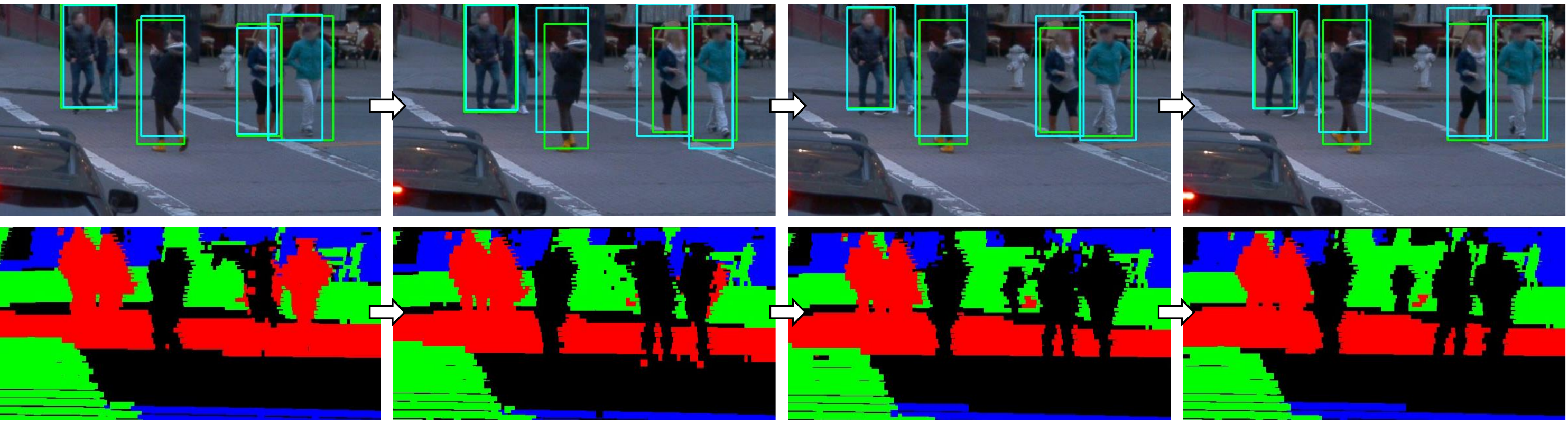}
    \caption {\small{\textbf{Multiple pedestrians tracking:} The top row shows that four pedestrians are detected in four consecutive frames and our method can track all of them. The green and cyan bounding boxes in the top row represent detected and tracked pedestrians, respectively. On the other hand, the bottom row shows the corresponding LiDAR scans where the relative distance of objects from the ego vehicle is shown by a jet color spectrum correlated to the 8-bit image scale ranging from the LiDAR's minimum to maximum value.
    }}
    \label{fig:Tracking_example1}
    \vspace{-10pt}
\end{figure*}
\begin{figure*}
    \centering
    \includegraphics[height=4.9cm]{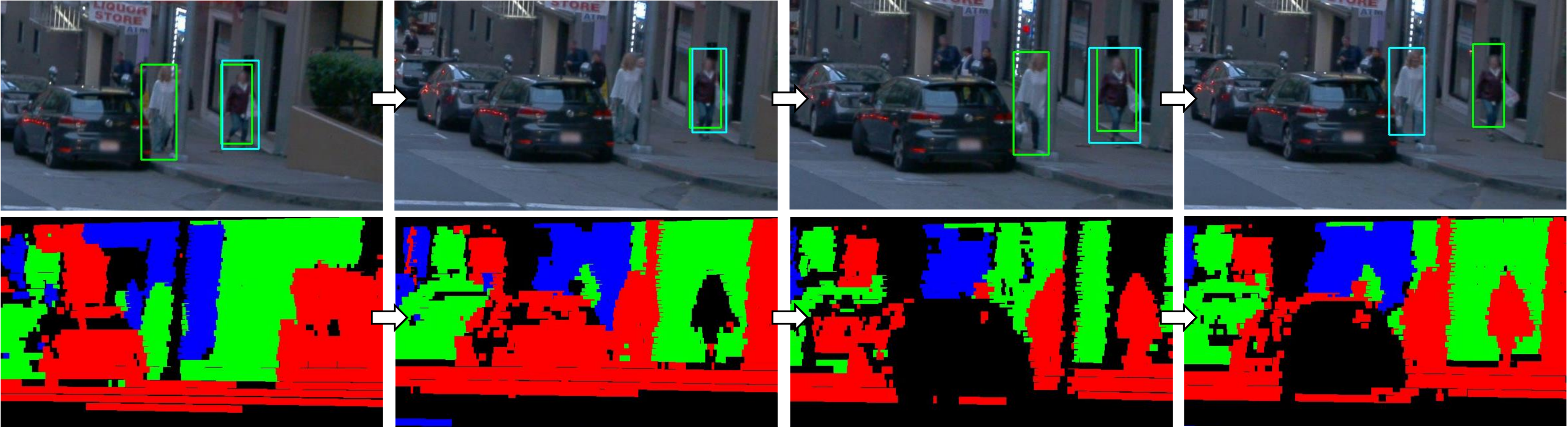}
    \caption {\small{\textbf{Robust pedestrians tracking:} In the top row, there are two pedestrians detected in the first frame and then in the second frame model failed to detect one of the pedestrians. However, in the next two consecutive frames that pedestrian is detected and tracked again which represents the robustness of our method. The green and cyan bounding boxes in the top row represent detected and tracked pedestrians, respectively. On the other hand, the bottom row shows the corresponding LiDAR scans where the relative distance of objects from the ego vehicle is shown by a jet color spectrum.}}
    \label{fig:Tracking_example2}
\end{figure*} 
 
\subsection{Performance Analysis}
We evaluate our proposed Fused-YOLO and the baseline YOLOv5 in terms of Miss Rate vs False Positive Per Image (FPPI) curve in Fig.~\ref{fig:Miss_rate_vs_fppi}. Testing on $456$ numbers of images from the Waymo dataset, the proposed Fused-YOLO shows the miss rate of $33.435\%$ whereas the YOLOv5 has the miss rate of $41.945\%$.
This is because fusion helps more for accurately detecting pedestrians in ill-lit conditions compare to only image-based pedestrian detection using a baseline detector. In low illumination conditions, the shape of pedestrians given by depth images entails useful features for pedestrian detection, which are more challenging to detect than camera images. Especially, we notice that the Fused-YOLO achieves significantly less miss rate in low illumination conditions in contrast to the baseline YOLOv5 model. Because the false detection increases when the bounding boxes from RGB image are naively fused with the bounding boxes of depth image. 
On the other hand, fusion without the Kalman filter exhibits a $38.602\%$ miss rate. On contrary, Fused-Yolo provides a systematic approach to reduce false detection and achieves the lowest miss rate.   

TABLE~\ref{tab:evaluation_metrics} shows that our proposed Fused-YOLO method has better accuracy, precision, and recall compared to the baseline YOLOv5. We observe that the YOLOv5 shows the accuracy of $0.546$, precision of $0.897$, and recall of $0.583$, whereas the Fused-YOLO has the accuracy of $0.609$, the precision of $0.900$ and, recall of $0.650$. The downside of Fused-YOLO is that it increases false detection when combining with the uncorrelated bounding boxes from depth images. One of the possible reasons is the fact that our Fused-YOLOv5 was not trained on depth images. Therefore, it struggles to accurately detect pedestrians from depth images.


We find that our tracking method performs very well even in the cases where there are some miss detections in sequential frames. 
Fig.~\ref{fig:Tracking_example1} illustrates the performance of our tracker on a sequence of images where four pedestrians are detected in four consecutive frames and our method can track all of them. The green and cyan bounding boxes in the top row represent detected and tracked pedestrians, respectively.
Fig.~\ref{fig:Tracking_example2} shows that there are two pedestrians detected in the first frame and then in the second frame model failed to detect one of the pedestrians. However, Fused-Yolo can track all two pedestrians by correlating the previous bounding boxes to the current estimation. Thanks to the Kalman filter which can track pedestrians even if the detector might fail to detect pedestrians on the current frame. Furthermore, additional detected bounding boxes from depth images help the Fused-YOLO to track the pedestrians robustly. Thus, our method is capable of reducing miss detection. We observe that the Fused-YOLO achieves negligible runtime performance overhead (28 FPS) in contrast to the baseline YOLOv5 model (30 FPS), resulting in real-time operation on the autonomous vehicle platform. 



\section{Conclusion}\label{Conclusion}
We developed a real-time accurate pedestrian detection and tracking framework by fusing camera and LiDAR sensor data. The developed framework is integrated with the Kalman filter to accurately and robustly detect and track multiple pedestrians.
The novelty of our framework lies in the adoption of the Kalman filter for both sensor fusion and tracking applications while minimizing the overall runtime overhead. Experimental results demonstrated the improvement over the baseline YOLOv5 model. Our fusion method outperforms YOLOv5 in terms of detection and tracking accuracy with a negligible amount of runtime overhead. The difference becomes even more pronounced in ill-lit conditions when pedestrians are hard to find in camera images. Finally, our implementation of the Kalman filter along with the optical flow algorithm reduced the detection miss rate and improved the overall performance. Future work includes extending our tracking method for pedestrians' behavior/intention analysis.
\section*{Acknowledgment}
The authors also thank the support from the North Carolina Department of Transportation (NCDOT) under the award numbers RP2019-28 and TCE2020-03, the National Science Foundation under award numbers 2018879 and 2000320, as well as  the support from SAE International and Ford Motor Company.

\bibliographystyle{config/IEEEtran/IEEEtran}  
\bibliography{ref1}

\end{document}